\documentclass{article}
\usepackage{iclr2027_conference,times}

\usepackage{amsmath,amsfonts,bm}

\def\eqref#1{equation~\ref{#1}}

\def\1{\bm{1}}

\DeclareMathAlphabet{\mathsfit}{\encodingdefault}{\sfdefault}{m}{sl}
\SetMathAlphabet{\mathsfit}{bold}{\encodingdefault}{\sfdefault}{bx}{n}

\renewcommand{\eqref}[1]{\textup{(\ref{#1})}} 
\usepackage{hyperref}
\usepackage{url}
\usepackage{amsmath,amssymb,mathtools,amsthm}
\usepackage{booktabs,multirow,adjustbox}
\usepackage{graphicx}
\usepackage{subcaption}
\usepackage{algorithm}
\usepackage{algpseudocode}
\usepackage{xspace}
\usepackage{tikz}
\usetikzlibrary{arrows.meta,positioning,calc,backgrounds}

\usepackage[most]{tcolorbox}
\usepackage{tabularx}
\usepackage{booktabs}
\usepackage{adjustbox}
\usepackage[table]{xcolor}

\newtcolorbox{promptbox}[1]{
    colback=gray!3,
    colframe=black!60,
    boxrule=0.6pt,
    arc=1mm,
    title=\textbf{#1},
    fonttitle=\small,
    fontupper=\small,
    breakable
}

\newcommand{\RELACE}{\textsc{RELACE}\xspace}

\title{RELACE: retrospective likelihood-based action credit estimation for long-horizon language agents}

\author{
Sayak Chakrabarti\thanks{Work done during an internship at Zoom Communications, Inc.} \\
Columbia University \\
sayaksc@gmail.com
\And
Sathish Reddy Indurthi\thanks{Project Mentor.} \\
Zoom Communications, Inc. \\
sathishreddy.indurthi@zoom.us
}

\iclrfinalcopy
\begin{document}
\maketitle 

\begin{abstract}

Group Relative Policy Optimization (GRPO) avoids a separate critic by estimating advantages from rollout groups. For multi-turn agents, however, trajectory-level supervision provides coarse, noisy credit: terminal rewards do not locate errors and can penalize useful actions alongside mistakes. Group-in-Group Policy Optimization (GiGPO) and subsequent methods refine supervision through state-conditioned comparisons, but their credit estimates remain sensitive to downstream decisions and outcomes. We introduce \RELACE, Retrospective Likelihood-based Action, a critic-free framework that integrates retrospective action assessment with state-conditioned advantage estimation. RELACE evaluates executed actions through teacher-forced likelihood scoring under both their original contexts and outcome-augmented contexts. Comparing these likelihoods yields a trajectory-normalized retrospective factor that captures outcome-dependent changes in action plausibility, rather than hindsight plausibility alone. We use this factor to reweight discounted task returns and construct local advantages by comparing weighted returns among actions from equivalent states within a task. This couples retrospective relevance with observed reward, producing fine-grained credit that complements trajectory-level GRPO supervision. Temporal smoothing and success-protecting masking further stabilize the local signal. RELACE requires neither auxiliary value nor reward models nor additional autoregressive rollouts for credit estimation. Experiments on ALFWorld and WebShop with Qwen2.5-1.5B-Instruct and Qwen2.5-7B-Instruct demonstrate substantial improvements over GRPO, GiGPO, and HCAPO. With the 1.5B model, RELACE achieves $96.35\%$ success on ALFWorld and $79.43\%$ on WebShop, surpassing GiGPO by $5.47$ and $5.60$ percentage points, respectively.
\end{abstract}

\begin{center}
    \small\textbf{Code:} \url{https://github.com/sayaksc/RELACE}
\end{center}

\section{Introduction}
\label{sec:introduction}

Large language models (LLMs) can act as autonomous agents, making sequential decisions in partially observable environments~\citep{yao2023react}. Tasks such as web navigation and simulated household interaction require coordinated actions to achieve user goals~\citep{yao2022webshop,shridhar2021alfworld}. Training agents for these tasks often involves sparse or delayed rewards, including terminal success signals, destabilizing training by obscuring intermediate actions' contributions~\citep{feng2025gigpo}. This highlights the temporal credit-assignment problem: an outcome evaluates the trajectory as a whole but provides limited guidance about individual decisions~\citep{harutyunyan2019hindsight}.

Group Relative Policy Optimization (GRPO) avoids a separate value critic by estimating relative advantages from samples for the same task~\citep{shao2024deepseekmath}. With outcome-level supervision, however, GRPO assigns the same scalar advantage to all actions within a trajectory. This coarse signal does not distinguish useful decisions from incidental or redundant actions, and a late error can obscure useful earlier behavior~\citep{feng2025gigpo}. The challenge is to obtain informative action-level credit while retaining critic-free simplicity.

Group-in-Group Policy Optimization (GiGPO) compares discounted downstream returns among decisions from matched environment states~\citep{feng2025gigpo}. Such comparisons account for the decision context, but their targets still reflect subsequent actions and transitions. Moreover, limited rollout sampling can make local estimates unreliable, assigning disproportionately large advantages to rare actions with favorable continuations~\citep{li2026ecpo}. State-level comparison supports local credit assignment, while leaving room for information about an action's relationship to the outcome.

Hindsight learning offers complementary information. Hindsight Experience Replay reuses achieved outcomes as alternative goals~\citep{andrychowicz2017hindsight}, whereas Hindsight Credit Assignment uses outcome-conditioned action distributions to inform credit allocation~\citep{harutyunyan2019hindsight}. For language agents, this perspective motivates a practical question: \emph{how does outcome information change the model's assessment of an executed action?} An action may score highly in hindsight because it was already plausible in its original context. Comparing both contexts can distinguish baseline plausibility from outcome sensitivity.

HCAPO applies hindsight-conditioned action scoring to long-horizon language-agent optimization~\citep{tan2026hcapo}. Although its theoretical formulation motivates a hindsight-to-original probability ratio, its practical estimator normalizes each hindsight score by the mean hindsight score along the trajectory. It does not explicitly evaluate each executed action's original-context denominator. This raises a design question: how should explicit original-versus-hindsight scoring be combined with state-conditioned comparisons of downstream returns?

We propose \RELACE, a critic-free framework that combines explicit retrospective scoring with state-conditioned advantage estimation. After termination, \RELACE augments each intermediate context with the observed terminal outcome. The frozen behavior policy scores the same executed action under original and outcome-conditioned contexts using teacher forcing. We form a ratio of these length-normalized likelihood-based scores, normalize the ratios within each trajectory, and clip them before reweighting discounted terminal returns. The resulting weights represent retrospective relevance rather than causal contribution.

\RELACE applies temporal smoothing to the weighted returns and computes local relative advantages within state-matched groups. Thus, outcome information enters the return targets \emph{before} local comparison, while a trajectory-level GRPO advantage retains supervision from overall task performance. Following HCAPO, we use temporal smoothing and an asymmetric mask that removes negative local corrections from successful trajectories~\citep{tan2026hcapo}. The environment reward is unchanged; \RELACE modifies advantage-estimation targets without a separate additive reward. Teacher-forced scoring adds likelihood evaluation without autoregressive rollouts or auxiliary value or reward models.

Our primary contributions are:
\begin{itemize}
    \item We develop an explicit original-versus-hindsight scoring mechanism for executed language actions. Ratios of length-normalized action scores yield trajectory-normalized, clipped weights for retrospective credit estimation using the behavior policy itself.

    \item We integrate these weights into a state-conditioned advantage estimator by reweighting discounted task returns before local group comparison. The resulting local signal complements trajectory-level GRPO supervision within critic-free optimization.

    \item We evaluate \RELACE on ALFWorld and WebShop using Qwen2.5-1.5B-Instruct and Qwen2.5-7B-Instruct. With the 1.5B model, \RELACE achieves the highest reported mean success rates among the compared methods: $96.35\%$ on ALFWorld and $79.43\%$ on WebShop. These exceed GiGPO by $5.47$ and $5.60$ percentage points, HCAPO by $9.35$ and $10.93$ points, and GraphGPO by $3.64$ and $0.78$ points, respectively.
   
\end{itemize}

\section{Related work}

\paragraph{Credit Assignment in Agentic Reinforcement Learning.}
Long-horizon language-agent tasks require learning from rewards that provide limited feedback about individual decisions~\citep{feng2025gigpo}. Actor--critic implementations of PPO estimate advantages using learned value functions, introducing additional model training and value-estimation error~\citep{schulman2017ppo}. Process supervision offers an alternative source of intermediate feedback through learned step-level evaluators, as demonstrated in mathematical reasoning~\citep{lightman2023verify}. Critic-free methods such as REINFORCE Leave-One-Out (RLOO) and Group Relative Policy Optimization (GRPO) instead construct advantages from sampled returns without fitting a separate value network~\citep{ahmadian2024rloo,shao2024deepseekmath}. When used with outcome-level supervision, however, a trajectory-level scalar advantage does not distinguish the roles of individual actions, motivating finer-grained credit estimators for multi-turn interaction~\citep{feng2025gigpo}.

\paragraph{State-Conditioned and Structured Credit Assignment.}
GiGPO constructs local relative advantages by grouping decisions with matching environment states and comparing their discounted downstream returns~\citep{feng2025gigpo}. HGPO addresses historical-context inconsistency within step-level groups by constructing a hierarchy of groups with different degrees of context agreement~\citep{hgpo2026}. GraphGPO aggregates sampled trajectories into a state-transition graph and assigns credit according to estimated progress toward task goals~\citep{cheng2026graphgpo}. ECPO focuses on the reliability of local comparisons, shrinking low-count action estimates and downweighting groups dominated by within-action noise~\citep{li2026ecpo}. These approaches refine the comparison structure, progress signal, or statistical reliability of step-level credit. \RELACE builds on state-conditioned grouping and investigates how outcome-conditioned action scores can inform the return targets used in those comparisons.

\paragraph{Exploration and Training Stability.}
Complementary work studies the learning dynamics of multi-turn agents. RAGEN examines training instability and the effects of rollout diversity, interaction granularity, and sampling frequency~\citep{wang2025ragen}. EPO regulates policy entropy across training phases to balance exploration and exploitation~\citep{xu2025epo}. EMPG modulates policy-gradient signals using step-wise uncertainty and final task outcomes, with an additional signal favoring predictable future interactions~\citep{wang2025empg}. These methods provide complementary perspectives on how exploration and update weighting affect learning from sparse feedback.

\paragraph{Hindsight Credit Assignment and Language Agents.}
Hindsight methods use information observed after an action to improve learning from collected experience. Hindsight Experience Replay (HER) relabels experience with achieved goals, enabling learning even when the intended goal is not reached~\citep{andrychowicz2017hindsight}. Hindsight Policy Gradients extends this idea to goal-conditioned policy-gradient estimation~\citep{rauber2019hpg}. Hindsight Credit Assignment (HCA) uses outcome-conditioned action distributions to estimate the relationship between earlier actions and subsequent outcomes~\citep{harutyunyan2019hindsight}. Building on HCA, COCOA estimates counterfactual contributions to rewards or reward-relevant outcome representations, addressing limitations of conditioning on detailed future states~\citep{meulemans2023cocoa}.

HCAPO brings hindsight-based credit assignment to long-horizon language agents through outcome-conditioned action scoring and a combination of trajectory-level and step-level advantages~\citep{tan2026hcapo}. Although its theoretical formulation motivates a hindsight-to-original probability ratio, its practical estimator normalizes each hindsight score by the mean hindsight score along the trajectory. \RELACE explicitly scores each executed action under both its original and outcome-conditioned contexts, then incorporates the resulting relevance weights into state-conditioned return comparisons. Its distinction lies in explicit original-context scoring, integration with local group baselines, teacher-forced evaluation, temporal smoothing, and protective masking. The complete estimator is presented in Section~\ref{sec:hegpo}.

\section{Retrospective likelihood-based action credit assignment}
\label{sec:hegpo}

We introduce \RELACE, a critic-free policy optimization algorithm for fine-grained credit assignment in partially observable multi-turn interactions. \RELACE combines three components: (i) an episode-level relative advantage capturing global trajectory quality, (ii) a retrospective likelihood ratio estimating intermediate actions' relevance to the outcome, and (iii) a state-conditioned baseline comparing diverging actions from matched environment states. The term \emph{retrospective} emphasizes that \RELACE measures how outcome information changes the policy's assessment of an executed action, rather than the outcome-conditioned score alone.

\subsection{Outcome-conditioned hindsight relabeling}
\label{sec:hindsight-relabeling}

For each task $x$, we sample $K$ trajectories from the behavior policy $\pi_{\theta_{\mathrm{old}}}$. Trajectory $\tau_i$ contains $T_i$ actions, receives terminal reward $R_i$, and ends with observed outcome $s_i^{\mathrm{final}}$. At step $t$, $s_{i,t}$ is the observable representation used for grouping, $c_{i,t}$ is the policy's interaction context, and $a_{i,t}=(y_{i,t,1},\ldots,y_{i,t,L_{i,t}})$ is the executed token sequence. The latent environment state need not be observable. Appendix~\ref{sec:preliminaries} gives the full setting and background.

Following hindsight learning~\citep{andrychowicz2017hindsight,harutyunyan2019hindsight}, we construct a retrospective context after termination, using the observed outcome for both successful and unsuccessful trajectories:
\begin{equation}
    c^{\mathrm{H}}_{i,t}
    =
    \operatorname{Relabel}
    \left(c_{i,t},s^{\mathrm{final}}_i\right).
    \label{eq:hegpo-hindsight-context}
\end{equation}

The outcome information is the final observation $s_i^{\mathrm{final}}$, used as the hindsight target. The operator $\operatorname{Relabel}$ constructs this prompt using the templates in Appendix~\ref{app:prompt_hindsight}, retaining the current observation and available interaction history.

We then evaluate the same executed action $a_{i,t}$ under both $c_{i,t}$ and $c^{\mathrm{H}}_{i,t}$, using the frozen behavior policy $\pi_{\theta_{\mathrm{old}}}$ and teacher forcing. No new action is sampled. Holding the executed action fixed measures the score change between the two contexts. The terminal outcome is used only for training-time credit estimation, never for rollout generation or evaluation.

\definecolor{stateblue}{RGB}{218,235,255}
\definecolor{stateblueborder}{RGB}{54,123,245}
\definecolor{stateorange}{RGB}{255,239,210}
\definecolor{stateorangeborder}{RGB}{255,145,38}
\definecolor{stategreen}{RGB}{220,244,218}
\definecolor{stategreenborder}{RGB}{69,160,72}
\definecolor{statered}{RGB}{255,225,225}
\definecolor{stateredborder}{RGB}{255,91,91}
\definecolor{statepurple}{RGB}{238,224,255}
\definecolor{statepurpleborder}{RGB}{153,82,255}
\definecolor{successfill}{RGB}{218,241,220}
\definecolor{successborder}{RGB}{62,140,70}
\definecolor{failurefill}{RGB}{255,220,220}
\definecolor{failureborder}{RGB}{255,80,80}
\definecolor{rowfill}{RGB}{248,250,252}
\definecolor{rowborder}{RGB}{210,220,232}
\definecolor{creditblue}{RGB}{0,78,180}
\definecolor{creditred}{RGB}{230,50,50}

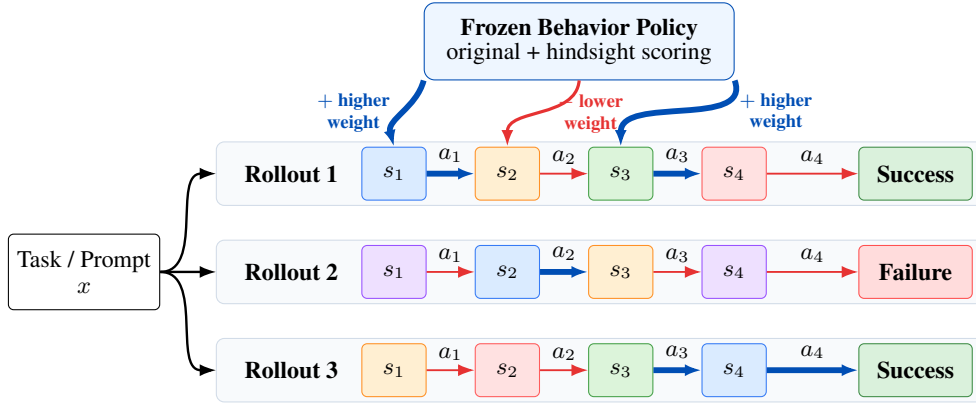
\begin{figure}[t]
\centering

\begin{tikzpicture}[
    font=\small,
    >={Latex[length=2.2mm,width=1.6mm]},
    state/.style={
        rounded corners=2pt,
        minimum width=8.5mm,
        minimum height=7mm,
        inner sep=1pt,
        font=\small
    },
    blueState/.style={
        state,
        fill=stateblue,
        draw=stateblueborder
    },
    orangeState/.style={
        state,
        fill=stateorange,
        draw=stateorangeborder
    },
    greenState/.style={
        state,
        fill=stategreen,
        draw=stategreenborder
    },
    redState/.style={
        state,
        fill=statered,
        draw=stateredborder
    },
    purpleState/.style={
        state,
        fill=statepurple,
        draw=statepurpleborder
    },
    outcome/.style={
        rounded corners=2pt,
        minimum height=7mm,
        minimum width=15mm,
        inner sep=2pt,
        font=\small\bfseries
    },
    success/.style={
        outcome,
        fill=successfill,
        draw=successborder
    },
    failure/.style={
        outcome,
        fill=failurefill,
        draw=failureborder
    },
    normalAction/.style={
        ->,
        draw=creditred,
        line width=0.7pt
    },
    highAction/.style={
        ->,
        draw=creditblue,
        line width=2.2pt
    },
    judgeHigh/.style={
        ->,
        draw=creditblue,
        line width=2.2pt
    },
    judgeLow/.style={
        ->,
        draw=creditred,
        line width=1.2pt
    }
]

\begin{scope}[on background layer]
    \filldraw[
        fill=rowfill,
        draw=rowborder,
        rounded corners=3pt
    ]
    (0.55,0.88) rectangle (10.70,1.72);

    \filldraw[
        fill=rowfill,
        draw=rowborder,
        rounded corners=3pt
    ]
    (0.55,-0.42) rectangle (10.70,0.42);

    \filldraw[
        fill=rowfill,
        draw=rowborder,
        rounded corners=3pt
    ]
    (0.55,-1.72) rectangle (10.70,-0.88);
\end{scope}

\node[
    draw,
    rounded corners=2pt,
    minimum width=1.5cm,
    minimum height=1cm,
    align=center,
    fill=white
] (task) at (-1.2,0)
{Task / Prompt\\$x$};

\node[font=\small\bfseries,anchor=west]
    (r1lab) at (0.80,1.30) {Rollout 1};

\node[font=\small\bfseries,anchor=west]
    (r2lab) at (0.80,0.00) {Rollout 2};

\node[font=\small\bfseries,anchor=west]
    (r3lab) at (0.80,-1.30) {Rollout 3};

\node[blueState]   (r1s1) at (2.90,1.30) {$s_1$};
\node[orangeState] (r1s2) at (4.40,1.30) {$s_2$};
\node[greenState]  (r1s3) at (5.90,1.30) {$s_3$};
\node[redState]    (r1s4) at (7.40,1.30) {$s_4$};
\node[success]     (r1out) at (9.80,1.30) {Success};

\draw[highAction]
    (r1s1) --
    node[above,text=black] {$a_1$}
    (r1s2);

\draw[normalAction]
    (r1s2) --
    node[above,text=black] {$a_2$}
    (r1s3);

\draw[highAction]
    (r1s3) --
    node[above,text=black] {$a_3$}
    (r1s4);

\draw[normalAction]
    (r1s4) --
    node[above,text=black] {$a_4$}
    (r1out);

\node[purpleState] (r2s1) at (2.90,0.00) {$s_1$};
\node[blueState]   (r2s2) at (4.40,0.00) {$s_2$};
\node[orangeState] (r2s3) at (5.90,0.00) {$s_3$};
\node[purpleState] (r2s4) at (7.40,0.00) {$s_4$};
\node[failure]     (r2out) at (9.80,0.00) {Failure};

\draw[normalAction]
    (r2s1) --
    node[above,text=black] {$a_1$}
    (r2s2);

\draw[highAction]
    (r2s2) --
    node[above,text=black] {$a_2$}
    (r2s3);

\draw[normalAction]
    (r2s3) --
    node[above,text=black] {$a_3$}
    (r2s4);

\draw[normalAction]
    (r2s4) --
    node[above,text=black] {$a_4$}
    (r2out);

\node[orangeState] (r3s1) at (2.90,-1.30) {$s_1$};
\node[redState]    (r3s2) at (4.40,-1.30) {$s_2$};
\node[greenState]  (r3s3) at (5.90,-1.30) {$s_3$};
\node[blueState]   (r3s4) at (7.40,-1.30) {$s_4$};
\node[success]     (r3out) at (9.80,-1.30) {Success};

\draw[normalAction]
    (r3s1) --
    node[above,text=black] {$a_1$}
    (r3s2);

\draw[normalAction]
    (r3s2) --
    node[above,text=black] {$a_2$}
    (r3s3);

\draw[highAction]
    (r3s3) --
    node[above,text=black] {$a_3$}
    (r3s4);

\draw[highAction]
    (r3s4) --
    node[above,text=black] {$a_4$}
    (r3out);

\draw[->,line width=0.9pt]
    (task.east)
    to[out=0,in=180]
    (0.55,1.30);

\draw[->,line width=0.9pt]
    (task.east) -- (0.55,0);

\draw[->,line width=0.9pt]
    (task.east)
    to[out=0,in=180]
    (0.55,-1.30);

\node[
    draw=creditblue,
    rounded corners=4pt,
    fill=stateblue!45,
    minimum width=4.1cm,
    minimum height=1.02cm,
    align=center
] (judge) at (5.35,3.05)
{
    \textbf{Frozen Behavior Policy}\\[-1pt]
    \footnotesize original + hindsight scoring
};

\draw[judgeHigh]
    (judge.south west)
    to[out=-105,in=105]
    node[
        pos=0.48,
        left=1pt,
        align=center,
        text=creditblue,
        font=\scriptsize\bfseries
    ]
    {$+$ higher\\weight}
    (r1s1.north);

\draw[judgeLow]
    (judge.south)
    to[out=-95,in=90]
    node[
        pos=0.50,
        right=2pt,
        align=center,
        text=creditred,
        font=\scriptsize\bfseries
    ]
    {$-$ lower\\weight}
    (r1s2.north);

\draw[judgeHigh]
    (judge.south east)
    to[out=-70,in=90]
    node[
        pos=0.48,
        right=15pt,
        align=center,
        text=creditblue,
        font=\scriptsize\bfseries
    ]
    {$+$ higher\\weight}
    (r1s3.north);

\end{tikzpicture}

\caption{
Overview of \RELACE.
Multiple trajectories are sampled for the same task, with colors indicating matched or similar environment states across rollouts. After the terminal outcome is observed, the frozen behavior policy scores the same executed actions under original and retrospective contexts using teacher forcing. The resulting hindsight reweighted returns are compared across matched states to construct the step-level credit signal. Arrow widths illustrate relative retrospective weights.}
\label{fig:relace_overview}

\end{figure}

\subsection{Retrospective credit reweighting via likelihood ratios}
\label{sec:hegpo-credit}
To estimate retrospective relevance, we compare likelihood-based scores for the same executed action under its original and retrospective contexts. For $L_{i,t}>0$, we use a temperature-scaled geometric mean of token probabilities to account for action length. The two scores are:
\begin{equation}
    p_{i,t} = \exp\left( \frac{1}{\tau_{\mathrm{temp}}L_{i,t}} \sum_{j=1}^{L_{i,t}} \log \pi_{\theta_{\mathrm{old}}} \left( y_{i,t,j} \mid c_{i,t},y_{i,t,<j} \right) \right),
    \label{eq:hegpo-original-prob}
\end{equation}
\begin{equation}
    p^{\mathrm{H}}_{i,t} = \exp\left( \frac{1}{\tau_{\mathrm{temp}}L_{i,t}} \sum_{j=1}^{L_{i,t}} \log \pi_{\theta_{\mathrm{old}}} \left( y_{i,t,j} \mid c^{\mathrm{H}}_{i,t},y_{i,t,<j} \right) \right),
    \label{eq:hegpo-hindsight-prob}
\end{equation}
where \(\tau_{\mathrm{temp}}>0\) controls score sharpness. These length-normalized scores are not normalized probabilities over variable-length actions, and this score temperature is distinct from decoding temperature. We refer to $p^{\mathrm{H}}_{i,t}$ as the \emph{retrospective likelihood score} of the executed action.

With numerical stabilizer $\epsilon>0$, we first compute the raw retrospective score ratio:
\begin{equation}
    u_{i,t} = \frac{p^{\mathrm{H}}_{i,t}}{p_{i,t}+\epsilon}.
    \label{eq:hegpo-raw-ratio}
\end{equation}

A ratio $u_{i,t}>1$ indicates a higher score under the retrospective context; its magnitude depends on the stabilizer when $p_{i,t}$ is small. To account for varying magnitudes across trajectories, we normalize the ratio by its trajectory mean, \(\bar{u}_i = \frac{1}{T_i} \sum_{k=1}^{T_i}u_{i,k}\), yielding a clipped relative relevance:
\begin{equation}
    \rho_{i,t} = \operatorname{clip} \left( \frac{u_{i,t}}{\bar{u}_i+\epsilon}, \rho_{\min}, \rho_{\max} \right),
    \label{eq:hegpo-normalized-ratio}
\end{equation}
where $0<\rho_{\min}\leq1\leq\rho_{\max}$. We distinguish raw ratio $u_{i,t}$ from normalized, clipped weight $\rho_{i,t}$ used in credit estimation. Writing $Q_{i,t}=\gamma^{T_i-t}R_i$ for the original discounted return, we obtain:
\begin{equation}
    Q^{\mathrm{H}}_{i,t} = \rho_{i,t}\gamma^{T_i-t}R_i.
    \label{eq:hegpo-hindsight-q}
\end{equation}
This reweights the return target without changing the environment reward. Before clipping and ignoring $\epsilon$, the weights have trajectory mean one; this does not in general preserve the sum of discounted return targets. In particular, if $R_i=0$, every $Q^{\mathrm H}_{i,t}$ is zero, regardless of the hindsight scores.

\subsection{State-conditioned grouping and asymmetric masking}
\label{sec:state-grouping}
To compare an action with sampled alternatives at a
decision node, \RELACE groups actions originating from matched
observable environment states across the rollout batch~\citep{feng2025gigpo}.
For a given task \(x\) and anchor state \(s\), the \emph{anchor state group}
is defined as
\[
\mathcal{G}_{x}(s)
=
\{(i,t)\mid x_i=x,\; s_{i,t}=s\},
\]
which contains all batch occurrences from the same observed state. 

State matching determines which decisions share a local baseline. Here,
$s_{i,t}$ denotes the observable representation used for grouping, whereas
$c_{i,t}$ denotes the full policy context used for likelihood scoring.
Matched observations need not identify the same latent state under partial
observability. The matching procedure appears in
Appendix~\ref{sec:appendix-training}.

Following HCAPO~\citep{tan2026hcapo}, we use adjacent-step smoothing to share the weighted return signal with the preceding action:
\begin{equation}
    \widetilde{Q}^{\mathrm{H}}_{i,t}
    =
    \alpha Q^{\mathrm{H}}_{i,t}
    +
    (1-\alpha)
    Q^{\mathrm{H}}_{i,\min(t+1,T_i)},
    \qquad
    \alpha \in [0,1].
    \label{eq:hegpo-smoothing}
\end{equation}

At $t=T_i$, this leaves the terminal target unchanged; $\alpha=1$ disables smoothing. Both the focal target and the local baseline use the same smoothed, weighted quantity. For a nonempty group, its mean is
\begin{equation}
\mu_V(x,s)
=
\frac{1}{|\mathcal{G}_x(s)|}
\sum_{(j,k)\in\mathcal{G}_x(s)}
\widetilde{Q}^{\mathrm H}_{j,k},
\end{equation}
and its sample standard deviation is
\begin{equation}
\sigma_V(x,s)
=
\sqrt{
\frac{1}{|\mathcal{G}_x(s)| -1}
\sum_{(j,k)\in\mathcal{G}_x(s)}
\left(
\widetilde{Q}^{\mathrm H}_{j,k}
-
\mu_V(x,s)
\right)^2
},
\end{equation}
when $|\mathcal{G}_x(s)| \geq 2$. Singleton groups receive zero local advantage without evaluating their sample standard deviation. Constant-target groups also yield zero local advantage. For groups of size at least two, the state-relative advantage is:
\begin{equation}
    A^{\mathrm{H}}_{i,t} = \frac{ \widetilde{Q}^{\mathrm{H}}_{i,t} - \mu_V(x_i,s_{i,t}) }{ \sigma_V(x_i,s_{i,t})+\epsilon }.
    \label{eq:hegpo-micro-advantage}
\end{equation}
A successful trajectory can receive a negative local advantage when its weighted target falls below its matched-group mean; finite-sample noise can also affect that comparison. Following HCAPO~\citep{tan2026hcapo}, we apply a conservative \emph{do-no-harm mask}. Let \(d_i \in \{0,1\}\) indicate trajectory success; the protected micro-advantage becomes:
\begin{equation}
    \widehat{A}^{\mathrm{H}}_{i,t} = 
    \begin{cases}
        0, & d_i=1 \land A^{\mathrm{H}}_{i,t}<0, \\
        A^{\mathrm{H}}_{i,t}, & \text{otherwise}.
    \end{cases}
    \label{eq:hegpo-no-harm}
\end{equation}
The mask removes every negative local correction on successful trajectories, whether or not caused by noise, and leaves the macro-advantage unchanged.

\subsection{Macro advantage and policy objective}
\RELACE also computes an episode-wise macro advantage \(A^{\mathrm{E}}_i\) within each task group to preserve preference for high-reward trajectories:
\begin{equation}
    A^{\mathrm{E}}_i = \frac{R_i-\mu_R(x_i)}{\sigma_R(x_i)+\epsilon}.
    \label{eq:hegpo-episode-advantage}
\end{equation}
The final advantage combines the episode-level and protected local terms:
\begin{equation}
    A^{\mathrm{RELACE}}_{i,t} = A^{\mathrm{E}}_i + \omega \widehat{A}^{\mathrm{H}}_{i,t},
    \label{eq:hegpo-final-advantage}
\end{equation}
where \(\omega \geq 0\) balances global and local preferences. Credit estimates are computed with the frozen behavior policy and detached during policy optimization. The scalar \(A^{\mathrm{RELACE}}_{i,t}\) is broadcast to all valid tokens of action \(a_{i,t}\). Let $r_{i,t,j}(\theta)=\pi_\theta(y_{i,t,j}\mid c_{i,t},y_{i,t,<j})/\pi_{\theta_{\mathrm{old}}}(y_{i,t,j}\mid c_{i,t},y_{i,t,<j})$ be the policy-update ratio, evaluated under the original context. This token-level ratio is distinct from the retrospective score ratio $u_{i,t}$. The policy is updated using the clipped PPO surrogate objective~\citep{schulman2017ppo}:
\begin{align}
    \mathcal{J}_{\mathrm{RELACE}}(\theta) = \mathbb{E} \Big[ \min\big(&r_{i,t,j}(\theta) A^{\mathrm{RELACE}}_{i,t}, \nonumber \\
    &\operatorname{clip} (r_{i,t,j}(\theta),1-\epsilon_{\mathrm{ppo}},1+\epsilon_{\mathrm{ppo}}) A^{\mathrm{RELACE}}_{i,t} \big) \Big],
    \label{eq:hegpo-objective}
\end{align}

Here, the expectation uses the task, trajectory, and valid-token aggregation specified in Eq.~\eqref{eq:prelim_grpo_objective}, replacing $A_i^E$ with $A^{\mathrm{RELACE}}_{i,t}$. We retain the KL regularization against the reference policy from that objective. The hindsight context is used for credit estimation, not as the policy-update input. Algorithm~\ref{alg:relace} in Appendix~\ref{app:algorithm} gives the training procedure.

\section{Experiments and analysis}

\subsection{Experimental setup}
\paragraph{Benchmarks and metrics.}
We evaluate \RELACE on two multi-turn environments with sparse outcome feedback: ALFWorld~\citep{shridhar2021alfworld} and WebShop~\citep{yao2022webshop}. ALFWorld requires household planning across six task categories: Pick, Clean, Heat, Cool, Examine, and Pick 2. We report overall and category-specific success rates. WebShop requires product search, attribute inspection, and purchases satisfying natural-language requests. We report mean task score (0--100) and success rate (SR), measuring complete task satisfaction. These distinguish partial from full completion. Episodes are capped at $50$ environment steps in ALFWorld and $15$ in WebShop.

\paragraph{Models and comparisons.}
We initialize policies from Qwen2.5-1.5B-Instruct and Qwen2.5-7B-Instruct~\citep{qwen2024}. The comparison includes prompting-based agents, ReAct and Reflexion~\citep{yao2023react,shinn2023reflexion}; actor--critic training with PPO~\citep{schulman2017ppo}; and critic-free methods, RLOO, GRPO, GiGPO, HCAPO, and GraphGPO~\citep{ahmadian2024rloo,shao2024deepseekmath,feng2025gigpo,tan2026hcapo,cheng2026graphgpo}. Table~\ref{tab:results} combines our \RELACE results with published baseline configurations: GRPO, GiGPO, and GraphGPO follow \citet{cheng2026graphgpo}, while HCAPO follows \citet{tan2026hcapo}. We separately compare \RELACE and GiGPO validation histories at shared checkpoints. Thus, the benchmark table and learning curves characterize task performance and training dynamics.

\paragraph{Training and reporting.}
Our implementation extends the \texttt{verl-agent} framework~\citep{feng2025gigpo}.\footnote{\url{https://github.com/langfengq/verl-agent}}
For the 1.5B experiments, each training batch contains $16$ prompts with $K=8$ rollouts per prompt. The actor learning rate is $10^{-6}$, and the KL regularization coefficient is $0.01$. \RELACE uses the retrospective weighting, temporal smoothing, and asymmetric masking defined in Section~\ref{sec:hegpo}, without an auxiliary value or reward model. Validation is logged every five training steps; Figure~\ref{fig:learning_curves} covers steps $0$--$200$. Hyperparameters, seed coverage, and aggregation details appear in Appendix~\ref{sec:appendix-training}; agent and hindsight prompts appear in Appendix~\ref{app:prompts}.

\subsection{Main results}
\RELACE achieves the highest reported mean on all three reported aggregate metrics in Table~\ref{tab:results}---ALFWorld overall success, WebShop score and success rate---at both model sizes. The improvements extend beyond trajectory-level GRPO to the listed step-level and hindsight-based baselines, showing that retrospective likelihood correction can complement structured local comparisons.

\paragraph{Strong performance with a 1.5B policy.}
On ALFWorld, \RELACE reaches $96.35\%$ overall success, exceeding GRPO, GiGPO, HCAPO, and GraphGPO by $18.49$, $5.47$, $9.35$, and $3.64$ percentage points, respectively. The subtask breakdown shows a large Cool gain: $96.49\%$ versus $85.26\%$ for GraphGPO and $81.46\%$ for GiGPO. \RELACE also reaches $100\%$ on Pick and the highest Examine success among the listed 1.5B methods. The overall advantage includes gains on tasks requiring coordinated actions, while Clean, Heat, and Pick 2 remain strengths of other methods.

On WebShop, the policy achieves a score of $90.90$ and a success rate of $79.43\%$. The success-rate gains over GRPO, GiGPO, and HCAPO are $8.08$, $5.60$, and $10.93$ percentage points. Relative to GraphGPO, \RELACE improves score by $1.61$ points and success by $0.78$ percentage points. Both gains indicate higher task rewards alongside more completed shopping tasks.

\paragraph{Benefits persist at 7B scale.}
With Qwen2.5-7B-Instruct, \RELACE reaches $97.66\%$ ALFWorld success, improving over GiGPO and GraphGPO by $3.39$ and $2.35$ percentage points. It achieves $100\%$ on Pick, Clean, and Pick 2, and the highest Cool and Examine means among these methods. On WebShop, \RELACE obtains a score of $90.34$ and $82.29\%$ success, exceeding GiGPO by $3.62$ score points and $3.91$ success percentage points, and GraphGPO by $3.40$ and $1.98$, respectively.

These results support the combination of original-context likelihood correction and state-conditioned return comparison across both policy sizes.

\begin{table*}[t]
\caption{Benchmark performance of \RELACE and the listed baseline configurations. WebShop Score is the mean task success rate (SR) on a 0--100 scale and all ALFWorld entries are success percentages. Baseline values follow \citet{cheng2026graphgpo} and \citet{tan2026hcapo}; Boldface and underlining indicate the highest and second-highest means within each model-size block, respectively.}

\label{tab:results}
\centering
\small
\begin{adjustbox}{max width=\textwidth}
\begin{tabular}{@{}l|cc|ccccccc@{}}
\toprule
\multirow{2}{*}{\textbf{Method}} & \multicolumn{2}{c|}{\textbf{WebShop}} & \multicolumn{7}{c}{\textbf{ALFWorld}}\\
\cmidrule(lr){2-3}\cmidrule(lr){4-10}
& Score & SR & Pick & Clean & Heat & Cool & Examine & Pick 2 & \textbf{Avg}\\
\midrule

\multicolumn{10}{c}{\textit{Closed-Source Models}} \\
GPT-5.6-Sol-High
 & $28.0$ & $25.3$
 & $\mathbf{98.7}$ & $\mathbf{76.6}$ & $\underline{85.7}$ & $\mathbf{83.6}$ & $\mathbf{89.8}$ & $\mathbf{80.2}$ & $\mathbf{86.0}$ \\
GPT-5.6-Sol-Medium
 & $26.7$ & $22.3$
 & $\underline{95.4}$ & $\underline{64.3}$ & $\mathbf{87.6}$ & $\underline{79.8}$ & $\underline{88.2}$ & $\underline{73.1}$ & $\underline{80.8}$ \\
GPT-4o
 & $31.8$ & $23.7$
 & $75.3$ & $60.2$ & $21.6$ & $31.2$ & $56.7$ & $49.8$ & $48.0$ \\
Gemini-2.5-Pro
 & $\mathbf{42.5}$ & $\mathbf{35.9}$
 & $92.8$ & $63.2$ & $26.6$ & $62.1$ & $69.0$ & $58.7$ & $60.3$ \\

\midrule
\multicolumn{10}{c}{\textit{Prompting (Qwen2.5-1.5B-Instruct)}}\\
ReAct & 40.1 & 11.3 & 17.4 & 20.5 & 7.7 & 15.7 & 6.2 & 2.0 & 12.8\\
Reflexion & 55.8 & 21.9 & 35.3 & 22.2 & 19.4 & 21.7 & 13.6 & 3.7 & 21.8\\

\midrule
\multicolumn{10}{c}{\textit{RL Training (Qwen2.5-1.5B-Instruct)}}\\
PPO & $73.8\pm3.0$ & $51.5\pm2.9$ & $64.8\pm3.5$ & $40.5\pm6.9$ & $46.4\pm4.0$ & $57.1\pm4.9$ & $60.6\pm6.6$ & $47.4\pm1.9$ & $54.4\pm3.1$\\
RLOO & $73.9\pm5.6$ & $52.1\pm6.7$ & $88.3\pm3.0$ & $52.8\pm8.6$ & $66.4\pm5.5$ & $71.0\pm5.9$ & $62.8\pm8.7$ & $56.9\pm4.7$ & $69.7\pm2.5$\\
GRPO & $84.73\pm0.49$ & $71.35\pm2.05$ & $82.89\pm3.62$ & $82.14\pm6.37$ & $77.78\pm4.54$ & $73.86\pm6.84$ & $78.57\pm0.00$ & $71.43\pm3.89$ & $77.86\pm1.33$\\
HCAPO & $83.8\pm0.7$ & $68.5\pm1.0$ & $88.6\pm7.0$ & $97.6\pm1.8$ & $90.7\pm6.9$ & $84.2\pm0.0$ & $75.0\pm0.0$ & $74.2\pm6.9$ & $87.0\pm4.1$\\
GiGPO & ${87.94\pm0.43}$ & $73.83\pm2.30$ & $\underline{98.81\pm1.68}$ & $95.16\pm3.89$ & ${94.44\pm0.00}$ & $81.46\pm0.56$ & $78.57\pm0.00$ & $\mathbf{93.65\pm5.94}$ & $90.88\pm0.97$\\
GraphGPO & $\underline{89.29\pm1.48}$ & $\underline{78.65\pm3.86}$ & $95.15\pm1.62$ & $\mathbf{100.0\pm0.00}$ & $\mathbf{96.30\pm2.61}$ & $\underline{85.26\pm2.58}$ & $\underline{85.71\pm5.83}$ & $\mathbf{93.65\pm2.24}$ & $\underline{92.71\pm1.32}$\\

\rowcolor{gray!20}
\textbf{\RELACE} & $\mathbf{90.90\pm1.04}$ & $\mathbf{79.43\pm2.74}$ & $\mathbf{100.00\pm0.00}$ & $\underline{98.41\pm2.75}$ & $\underline{94.45\pm9.61}$ & $\mathbf{96.49\pm6.08}$ & $\mathbf{87.46\pm11.38}$ & $\underline{93.33\pm3.12}$ & $\mathbf{96.35\pm0.45}$\\

\midrule
\multicolumn{10}{c}{\textit{RL Training (Qwen2.5-7B-Instruct)}}\\
GRPO & $84.31\pm1.27$ & $75.00\pm2.78$ & $88.98\pm5.30$ & $91.98\pm4.43$ & $90.74\pm5.24$ & $77.89\pm4.58$ & $78.57\pm0.00$ & $71.43\pm3.89$ & $83.33\pm2.05$\\
HCAPO & $85.1\pm1.3$ & $73.8\pm2.8$ & $\underline{99.10\pm1.30}$ & $\underline{97.30\pm1.90}$ & $81.80\pm8.80$ & ${90.80\pm6.60}$ & $90.30\pm2.00$ & $81.90\pm10.00$ & $91.40\pm2.30$\\
GiGPO & $86.72\pm1.44$ & $78.38\pm1.94$ & $97.53\pm1.75$ & $\mathbf{100.0\pm0.00}$ & $\mathbf{94.44\pm0.00}$ & $83.98\pm1.32$ & $90.48\pm6.73$ & $\mathbf{100.0\pm0.00}$ & $94.27\pm1.33$\\
GraphGPO & $\underline{86.94\pm0.68}$ & $\underline{80.31\pm1.33}$ & $\mathbf{100.0\pm0.0}$ & $\mathbf{100.0\pm0.0}$ & $\mathbf{94.44\pm0.0}$ & $\underline{91.40\pm1.5}$ & $\underline{92.86\pm5.8}$ & $\underline{92.06\pm2.2}$ & $\underline{95.31\pm1.10}$\\

\rowcolor{gray!20}
\textbf{\RELACE} & $\mathbf{90.34\pm0.80}$ & $\mathbf{82.29\pm1.80}$ & $\mathbf{100.00\pm0.00}$ & $\mathbf{100.00\pm0.00}$ & $\underline{94.11\pm5.37}$ & $\mathbf{94.71\pm5.03}$ & $\mathbf{94.44\pm9.62}$ & $\mathbf{100.00\pm0.00}$ & $\mathbf{97.66\pm0.78}$\\

\bottomrule
\end{tabular}\end{adjustbox}
\end{table*}

\subsection{Learning dynamics}
\label{sec:learning-dynamics}
\begin{figure}[t]
    \centering
    \begin{subfigure}[b]{0.48\textwidth}
        \includegraphics[width=\textwidth]{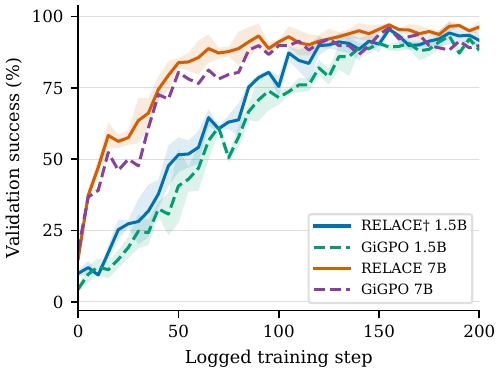}
        \caption{ALFWorld Success Rate}
    \end{subfigure}
    \hfill
    \begin{subfigure}[b]{0.48\textwidth}
        \includegraphics[width=\textwidth]{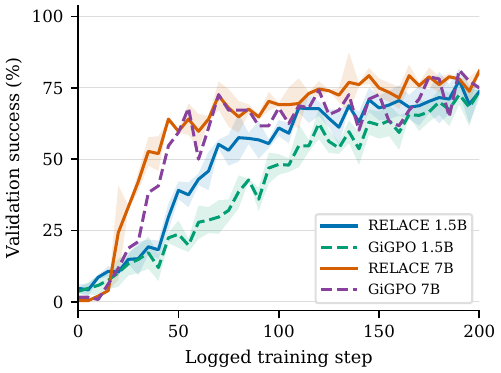}
        \caption{WebShop Success Rate}
    \end{subfigure}
    \caption{Validation success over steps $0$--$200$ for \RELACE (solid) and GiGPO (dashed), using Qwen2.5-1.5B/7B-Instruct. Lines show seed means; shading shows one sample standard deviation, omitted for single-seed baselines. Seed coverage and the $\dagger$ run-assignment note appear in Appendix~\ref{sec:appendix-training}.}
    \label{fig:learning_curves}
\end{figure}

Figure~\ref{fig:learning_curves} compares validation success over the first $200$ logged training steps. The curves compare \RELACE's learning progress with GiGPO, complementing the benchmark results in Table~\ref{tab:results}.

The 1.5B curves separate most during early and intermediate training. On ALFWorld at step $50$, \RELACE reaches $51.56\%$ success versus $40.63\%$ for GiGPO, a $10.94$-point advantage. On WebShop, the gap is larger: $39.06\%$ versus $23.70\%$, or $15.36$ points. The WebShop advantage remains $12.76$ points at step $100$ ($60.94\%$ versus $48.18\%$). The curves show faster early learning and higher success for RELACE. RELACE's computational overhead appears in Appendix \ref{sec:appendix-training}.

\subsection{Ablation study}
We examine two retrospective-estimator choices: correcting hindsight scores using the original action likelihood, and applying the resulting weights to the local comparison group. Table~\ref{tab:webshop_component_ablation} reports four WebShop variants. Using $Q_{i,t}=\gamma^{T_i-t}R_i$, GiGPO compares unweighted returns $Q$. Hindsight-score weighting uses $p^H Q$ for both the focal action and its group baseline, without the original-context denominator. Focal-only ratio weighting uses $\rho Q$ for the focal action but retains a baseline constructed from $Q$. Full \RELACE uses $\rho Q$ on both sides, with the smoothing and normalization defined in Section~\ref{sec:state-grouping}.

\begin{table}[t]
\centering
\small
\caption{Component comparison on WebShop. Columns show the focal target and group-baseline input before smoothing.}
\label{tab:webshop_component_ablation}
\begin{tabular}{lccc}
\toprule
Variant & Focal target & Baseline input & Reported SR (\%) \\
\midrule
GiGPO & $Q$ & $Q$ & $73.83$ \\
Hindsight-score weighting & $p^H Q$ & $p^H Q$ & $75.78$ \\
Focal-only ratio weighting & $\rho Q$ & $Q$ & $60.94^{\ast}$ \\
\RELACE & $\rho Q$ & $\rho Q$ & $\mathbf{79.43}$ \\
\bottomrule
\end{tabular}
\end{table}

\paragraph{Original-context correction.}
Hindsight-score weighting reaches $75.78\%$, compared with $73.83\%$ for GiGPO. Full \RELACE reaches $79.43\%$, a further $3.65$-point increase. This ordering supports comparing an action's hindsight and original plausibility: a high hindsight score may reflect an already probable action.

\paragraph{Consistency of the local baseline.}
The focal-only variant peaks at $60.94\%$, indicating that focal-return weighting alone is insufficient here. An unweighted baseline compares differently transformed quantities. In \RELACE, both focal targets and group statistics use the same hindsight-weighted, smoothed returns. The ordering supports this design motivation.

\subsection{Dynamics of retrospective scoring}
\label{sec:hindsight-score-dynamics}

\begin{figure}[t]
    \centering
    \includegraphics[width=\linewidth]{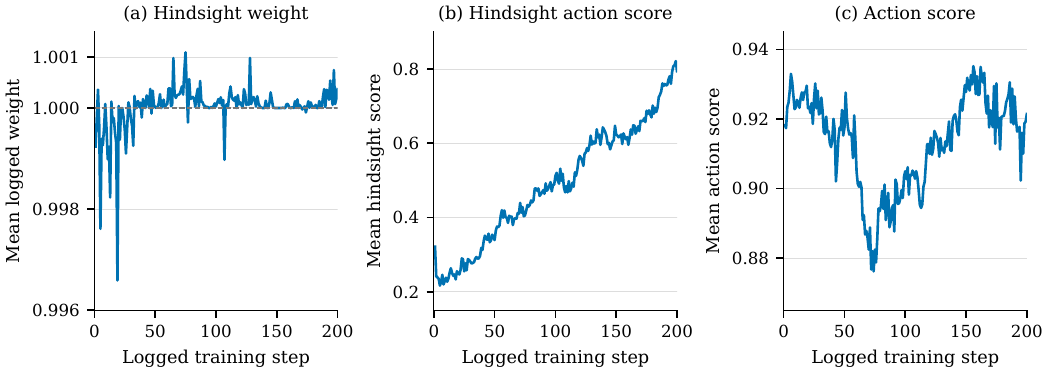}
    \caption{RELACE scoring dynamics on WebShop (Qwen2.5-1.5B-Instruct, steps $1$--$200$): (a) hindsight weight, with a dashed reference at one; (b) hindsight action score; and (c) action score.}
    \label{fig:hindsight-score-dynamics}
\end{figure}

Complementing validation, we examine $200$ training steps of Qwen2.5-1.5B-Instruct on WebShop. Figure~\ref{fig:hindsight-score-dynamics} tracks three run diagnostics: mean retrospective weight, mean hindsight score, and action score. These curves characterize scoring during training.

\paragraph{Stability of the mean retrospective weight.}
The logged mean weight stays near one, ranging from $0.9966$ to $1.0011$. Its temporal standard deviation falls from $6.84\times10^{-4}$ over steps $1$--$50$ to $1.58\times10^{-4}$ over steps $151$--$200$, a reduction of about $77\%$. This indicates greater late-stage stability of the batch mean. 
\paragraph{Increasing hindsight scores.}
The mean hindsight score rises from $0.241$ over steps $1$--$10$ to $0.785$ over steps $191$--$200$. Successive $50$-step averages are $0.279$, $0.434$, $0.563$, and $0.687$, showing an upward trend despite batch fluctuations. Thus, later actions score higher in outcome-conditioned contexts, consistent with greater compatibility with the final-observation hindsight targets.

\paragraph{Non-monotonic action-score evolution.}
The action score follows a different trajectory. Its $50$-step average decreases from $0.921$ over steps $1$--$50$ to $0.898$ over steps $51$--$100$, then recovers to $0.911$ and $0.924$ in the next two windows. Hindsight scores rise while action scores decline and recover, rather than both increasing uniformly. These patterns suggest changing action preferences during training.

\section{Conclusion}

Sparse terminal rewards provide limited credit-assignment guidance in long-horizon tasks. We introduced \RELACE, a critic-free framework that uses teacher forcing to compare executed actions under original and outcome-conditioned contexts. The relevance weights refine discounted returns before state-conditioned comparison, combining local credit with GRPO supervision without auxiliary models or additional autoregressive rollouts.

Experiments on ALFWorld and WebShop show higher reported aggregate performance than compared baselines at 1.5B and 7B scales, with early gains over GiGPO at 1.5B. Training diagnostics show increasing hindsight scores and a more stable mean retrospective weight, supporting learning from sparse outcomes.

These scores represent retrospective relevance rather than causal contribution. Future work will examine their relationship to action utility, improve partially observable state matching, and evaluate generalization across environments.

\subsection*{AI use statement}
Generative AI tools assisted with sentence-level editing, literature lookup and citation formatting. Also assisted in implementing  code experimentation, and LaTeX compilation checks. The authors remain responsible for the final text, analyses, and claims.



\newpage
\appendix

\section{Preliminaries}
\label{sec:preliminaries}

\paragraph{Markov decision processes.}
A Markov decision process (MDP) is defined by
$\mathcal{M}=(\mathcal{S},\mathcal{A},P,R,\gamma)$, where
$\mathcal{S}$ is the state space, $\mathcal{A}$ is the action space,
$P(s'\mid s,a)$ is the transition kernel, $R$ is the reward function,
and $\gamma\in[0,1]$ is the discount factor.
A policy $\pi(a\mid s)$ selects an action conditioned on the current
state, and the objective is to maximize expected discounted return.

\paragraph{Partial observability.}
Long-horizon language agents generally do not have direct access to the
underlying environment state. We therefore model interaction as a
partially observable Markov decision process (POMDP),
\[
\mathcal{M}
=
(\mathcal{Z},\mathcal{A},\mathcal{O},P,\Omega,R,\gamma),
\]
where $\mathcal{Z}$ denotes latent states, $\mathcal{O}$ observations,
and $\Omega(o\mid z)$ the observation kernel.
Rather than conditioning only on the current latent state, the policy
acts from its available interaction history.
For language agents, we represent this information by a context
$c_t$, containing the task specification, previous observations and
actions, and the current observation.

\paragraph{Language-agent setting.}
For each task $x$, we sample a group of $K$ trajectories from the
behavior policy $\pi_{\theta_{\mathrm{old}}}$.
Trajectory
\[
\tau_i =
(s_{i,1},a_{i,1},\ldots,s_{i,T_i},a_{i,T_i},
s_i^{\mathrm{final}})
\]
contains $T_i$ actions, terminates in observed outcome
$s_i^{\mathrm{final}}$, and receives terminal reward
$R_i=R(\tau_i)$.
At turn $t$, the context $c_{i,t}$ contains the task and observed
interaction history, while the action $a_{i,t}$ is a variable-length
token sequence
\[
a_{i,t}=(y_{i,t,1},\ldots,y_{i,t,L_{i,t}}).
\]
We use $s_{i,t}$ to denote the observable environment representation
used for state-conditioned grouping. Under partial observability,
identical observed states need not correspond to identical latent
environment states.

\paragraph{Group Relative Policy Optimization.}
GRPO~\citep{shao2024deepseekmath} avoids training a separate value
critic by constructing relative advantages from groups of rollouts
sampled for the same task. For trajectory $\tau_i$, the episode-level
advantage is
\begin{equation}
A_i^{E}
=
\frac{R_i-\mu_R(x)}
{\sigma_R(x)+\epsilon},
\qquad
r_{i,t,j}(\theta)
=
\frac{
\pi_\theta(y_{i,t,j}\mid c_{i,t},y_{i,t,<j})
}{
\pi_{\theta_{\mathrm{old}}}
(y_{i,t,j}\mid c_{i,t},y_{i,t,<j})
},
\label{eq:prelim_grpo_advantage}
\end{equation}
where $\mu_R(x)$ and $\sigma_R(x)$ are reward statistics computed over
the rollout group for task $x$, and $\epsilon>0$ ensures numerical
stability. GRPO optimizes the clipped surrogate objective
\begin{equation}
\begin{aligned}
\mathcal{J}_{\mathrm{GRPO}}(\theta)
&=
\mathbb{E}\!\left[
\frac{1}{K}
\sum_{i=1}^{K}
\frac{1}{N_i}
\sum_{t,j}
\left\{
\ell\!\left(
r_{i,t,j}(\theta),A_i^E
\right)
-
\beta D_{i,t,j}
\right\}
\right],\\
\ell(r,A)
&=
\min\!\left\{
rA,\,
\operatorname{clip}
(r,1-\epsilon_{\mathrm{ppo}},1+\epsilon_{\mathrm{ppo}})A
\right\},
\end{aligned}
\label{eq:prelim_grpo_objective}
\end{equation}
where $N_i=\sum_t L_{i,t}$ and $D_{i,t,j}$ denotes the token-level KL
penalty relative to a reference policy. The expectation is over tasks
and rollout groups, while $\beta$ and $\epsilon_{\mathrm{ppo}}$
control KL regularization and PPO clipping, respectively.

When only a terminal reward is available, $A_i^E$ is shared by every
action in trajectory $\tau_i$. GRPO therefore distinguishes trajectories
by overall outcome but does not by itself distinguish the contributions
of individual intermediate actions.

\paragraph{Hindsight credit assignment.}
Hindsight Credit Assignment (HCA)~\citep{harutyunyan2019hindsight}
uses future outcomes to retrospectively assess earlier actions.
For a Markov state $s$, define the hindsight action posterior
\[
h_k^\pi(a\mid s,s')
=
\Pr_\pi(A_0=a\mid S_0=s,S_k=s').
\]
For $\pi(a\mid s)>0$, HCA gives the action-value identity
\begin{equation}
Q^\pi(s,a)
=
r(s,a)
+
\mathbb{E}_{\tau\sim\pi(\cdot\mid s)}
\!\left[
\sum_{k\geq 1}
\gamma^k
\frac{
h_k^\pi(a\mid s,S_k)
}{
\pi(a\mid s)
}
r_k
\right],
\label{eq:prelim_hca}
\end{equation}
where $r(s,a)$ is the expected immediate reward and $r_k$ is a future
reward. The expectation is over trajectories beginning from $s$ with
$A_0\sim\pi(\cdot\mid s)$, rather than trajectories conditioned on
$A_0=a$.

For sparse terminal rewards, HCAPO~\citep{tan2026hcapo} motivates the
step-level quantity
\begin{equation}
Q^H_{i,t}
=
\frac{
h^\pi
(a_{i,t}\mid c_{i,t},s_i^{\mathrm{final}})
}{
\pi(a_{i,t}\mid c_{i,t})
}
\gamma^{T_i-t}R_i,
\label{eq:prelim_terminal_hindsight}
\end{equation}
with $\pi=\pi_{\theta_{\mathrm{old}}}$ and the hindsight distribution
conditioned on the observed terminal outcome.
In practical language-agent implementations, the exact hindsight
posterior is unavailable and must be approximated using policy
likelihoods under outcome-conditioned contexts.

The likelihood-based scores used by \RELACE are therefore practical
credit proxies rather than exact HCA posteriors; the HCA identity above
does not imply that the resulting estimator is unbiased.

\section{Full RELACE algorithm}
\label{app:algorithm}

Algorithm~\ref{alg:relace} summarizes the complete RELACE training procedure. The algorithm first samples grouped trajectories using the frozen behavior policy, retrospectively scores each executed action under its original and outcome-conditioned contexts, constructs trajectory-normalized likelihood-ratio weights, and uses these weights to form state-conditioned local advantages. These local advantages are then combined with the trajectory-level GRPO advantage for policy
optimization.

\begin{algorithm}[h]
\caption{Retrospective Likelihood-based Action Credit Estimation (RELACE)}
\label{alg:relace}
\begin{algorithmic}[1]
\Require Policy \(\pi_{\theta_{\mathrm{old}}}\), task batch \(\mathcal{X}\), group size \(K\), discount \(\gamma\), micro-advantage weight \(\omega\), smoothing coefficient \(\alpha\), ratio bounds \(\rho_{\min},\rho_{\max}\), and numerical constant \(\epsilon\)
\For{each task \(x\in\mathcal{X}\)}
    \State Sample \(K\) trajectories \(\{\tau_i\}_{i=1}^{K}\sim\pi_{\theta_{\mathrm{old}}}\)
    \State Record states, actions, final states, returns, and success labels
    \For{each trajectory \(\tau_i\)}
        \For{\(t=1,\ldots,T_i\)}
            \State Construct outcome-conditioned context \(c^{\mathrm{H}}_{i,t} \gets \operatorname{Relabel}(c_{i,t},s^{\mathrm{final}}_i)\)
            \State Teacher-force \(a_{i,t}\) under \(c_{i,t}\) and \(c^{\mathrm{H}}_{i,t}\)
            \State Compute \(p_{i,t}\) and \(p^{\mathrm{H}}_{i,t}\) using Eqs.~\eqref{eq:hegpo-original-prob}--\eqref{eq:hegpo-hindsight-prob}
            \State \(u_{i,t}\gets p^{\mathrm{H}}_{i,t}/(p_{i,t}+\epsilon)\)
        \EndFor
        \State \(\bar{u}_i\gets T_i^{-1}\sum_{t=1}^{T_i}u_{i,t}\)
        \For{\(t=1,\ldots,T_i\)}
            \State \(\rho_{i,t} \gets \operatorname{clip} (u_{i,t}/(\bar{u}_i+\epsilon), \rho_{\min},\rho_{\max})\)
            \State \(Q^{\mathrm{H}}_{i,t} \gets \rho_{i,t}\gamma^{T_i-t}R_i\)
        \EndFor
        \State Optionally smooth \(Q^{\mathrm{H}}_{i,t}\) using Eq.~\eqref{eq:hegpo-smoothing}
    \EndFor
    \State Group valid states within each task using the state-matching rule in Section~\ref{sec:state-grouping}
    \State Compute \(\mu_V(x,s)\) and \(\sigma_V(x,s)\)
    \State Compute state-relative retrospective likelihood-based advantages \(A^{\mathrm{H}}_{i,t}\)
    \State Apply the do-no-harm mask to obtain \(\widehat{A}^{\mathrm{H}}_{i,t}\)
    \State Compute episode-wise advantages \(A^{\mathrm{E}}_i\)
    \State Combine advantages: \(A^{\mathrm{RELACE}}_{i,t} \gets A^{\mathrm{E}}_i+\omega\widehat{A}^{\mathrm{H}}_{i,t}\)
\EndFor
\State Update \(\theta\) by maximizing \(\mathcal{J}_{\mathrm{RELACE}}(\theta)\)
\end{algorithmic}
\end{algorithm}

\section{Experimental details and hyperparameters}
\label{sec:appendix-training}

\subsection{Training configuration.}
For the 1.5B experiments, we use a training batch size of $16$ with
$K=8$ environment rollouts per prompt. The actor learning rate is
$10^{-6}$ and the KL regularization coefficient is $0.01$. KL shaping
of the environment reward is disabled. Rollouts are generated using
vLLM with sampling temperature $1.0$, while validation uses temperature
$0.4$ and a validation batch size of $128$. We evaluate the policy every
five training steps. Episodes are limited to $50$ environment steps in
ALFWorld and $15$ steps in WebShop.

For \RELACE, we set the temporal smoothing coefficient to
$\alpha=0.5$, the micro-advantage weight to $\omega=1$, and clip the
normalized retrospective likelihood ratio to
$[\rho_{\min},\rho_{\max}]=[0.8,1.2]$. The likelihood-score temperature
is $1$. Temporal smoothing and the success-protecting mask are enabled.
The discount factor is $\gamma=0.6$ for \RELACE and $\gamma=0.95$ for
the GiGPO baseline \cite{feng2025gigpo}. The PPO mini-batch size is $64$ for all experiments except for for WebShop 7B, where the mini-batch size is $128$.

\paragraph{State matching.}
The exact normalized equality in Section~\ref{sec:state-grouping} defines
idealized anchor groups. In practice, we construct state-conditioned groups
within each task using string-similarity matching with threshold $0.95$.
Exact matching can produce sparse groups when observable representations
differ textually, while approximate matching risks grouping distinct
decision situations.

\subsection{computational overhead analysis}

Following the stage-wise analysis of HCAPO~\citep[Section~6.5]{tan2026hcapo}, we profile \RELACE on WebShop with Qwen2.5-1.5B-Instruct on one node with eight NVIDIA H100 80GB GPUs. The run uses $16$ prompts and $8$ rollouts per prompt, an actor PPO mini-batch size of $64$, and a maximum of $15$ environment steps. We average the first $200$ timed training iterations (logged steps $1$--$200$), retaining initial iterations. This dedicated timing run is separate from the validation histories in Figure~\ref{fig:learning_curves}.

\begin{figure}[htbp]
\centering
\includegraphics[width=0.88\textwidth]{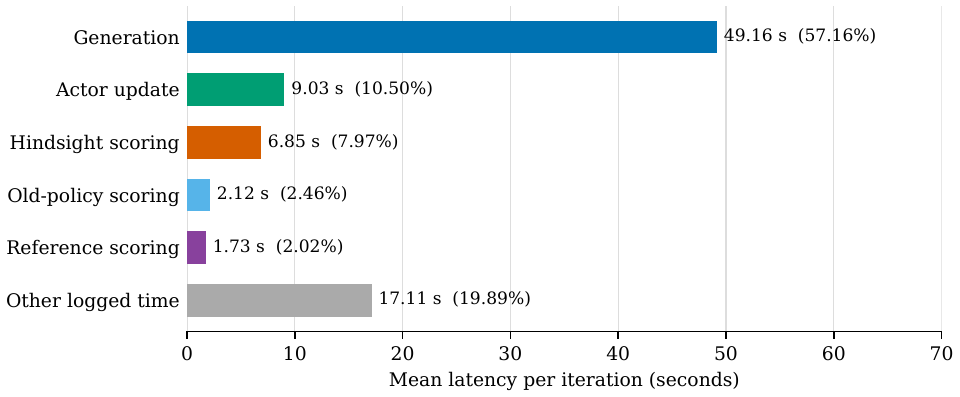}
\caption{Mean training latency on WebShop over logged steps $1$--$200$ of one timing run. Labels show seconds per iteration and the share of full logged iteration time. Hindsight scoring (orange) averages $6.85$\,s, or $7.97\%$ of the full iteration. ``Other logged time'' includes periodic evaluation, checkpointing, and residual work.}
\label{fig:timing-breakdown}
\end{figure}

\paragraph{Cost of retrospective scoring.}
Figure~\ref{fig:timing-breakdown} shows that generation remains the dominant stage at $49.16$\,s per iteration, followed by actor updates at $9.03$\,s. Hindsight scoring averages $6.85$\,s, accounting for $7.97\%$ of the full logged iteration time of $86.01$\,s. Restricting the denominator to the five major stages---generation, actor update, hindsight scoring, old-policy scoring, and reference scoring---gives $68.90$\,s and a hindsight share of $9.94\%$. The remaining $17.11$\,s comprises other logged work, including advantage and reward processing, periodic evaluation, and checkpoint saving.

The relatively small scoring share is consistent with the computation used by \RELACE: teacher forcing evaluates tokens from existing trajectories without additional autoregressive action generation or environment interaction. In this profile, the hindsight pass takes approximately $14\%$ of the generation time, making retrospective scoring a modest component of the measured training pipeline. These measurements quantify the recorded scoring stage; incremental end-to-end slowdown relative to GiGPO would require a matched run with hindsight scoring disabled. Timing aggregation details are provided in Appendix~\ref{sec:appendix-training}.

\section{Prompt templates}
\label{app:prompts}

For completeness and reproducibility, we provide the prompt templates used
for policy interaction in ALFWorld and WebShop, together with the
outcome-conditioned prompt used for retrospective likelihood evaluation.
Variables enclosed in braces are instantiated from the current environment
state and interaction history.

\subsection{ALFWorld agent prompt}
\label{app:prompt_alfworld}

\begin{promptbox}{Prompt Template for ALFWorld}
You are an expert autonomous agent operating in the ALFRED Embodied Environment.

Your task is to: \texttt{\{task\_description\}}.

Prior to this step, you have already taken \texttt{\{step\_count\}} step(s). Below are the most recent \texttt{\{history\_length\}} observations and the corresponding actions you took: \texttt{\{action\_history\}}

You are now at step \texttt{\{current\_step\}} and your current observation is: \texttt{\{current\_observation\}}.

Your admissible actions of the current situation are:
\[
\texttt{\{available\_actions\}}
\]

Now it's your turn to take one action for the current step.

You should first reason step-by-step about the current situation, then think carefully which admissible action best advances the shopping goal. This reasoning process MUST be enclosed within \texttt{<think>} \texttt{</think>} tags.

Once you've finished your reasoning, you should choose an admissible action for the current step and present it within \texttt{<action>} \texttt{</action>} tags.
\end{promptbox}

\subsection{WebShop agent prompt}
\label{app:prompt_webshop}

\begin{promptbox}{Prompt Template for WebShop}
You are an expert autonomous agent operating in the WebShop e-commerce environment.

Your task is to: \texttt{\{task\_description\}}.

Prior to this step, you have already taken \texttt{\{step\_count\}} step(s). Below are the most recent \texttt{\{history\_length\}} observations and the corresponding actions you took: \texttt{\{action\_history\}}

You are now at step \texttt{\{current\_step\}} and your current observation is: \texttt{\{current\_observation\}}.

Your admissible actions of the current situation are:
\[
\texttt{\{available\_actions\}}
\]

Now it's your turn to take one action for the current step.

You should first reason step-by-step about the current situation, then think carefully which admissible action best advances the shopping goal. This reasoning process MUST be enclosed within \texttt{<think>} \texttt{</think>} tags.

Once you've finished your reasoning, you should choose an admissible action for the current step and present it within \texttt{<action>} \texttt{</action>} tags.
\end{promptbox}

\subsection{Outcome-conditioned hindsight prompt}
\label{app:prompt_hindsight}

\RELACE performs retrospective scoring after the trajectory terminates.
For each executed action $a_{i,t}$, we retain the original decision context
$c_{i,t}$ and augment it with the observed terminal outcome
$s^{\mathrm{final}}_i$. The executed action itself is not regenerated;
instead, it is scored using teacher forcing under both contexts.

\begin{promptbox}{Hindsight Prompt Template for ALFWorld}
You are an expert agent operating in the ALFRED Embodied Environment.

Your task is to: Navigate the environment to reach the following exact observation:

\texttt{[Observation]: \{final\_obs\}}

Prior to this step, you have already taken \texttt{\{step\_count\}} step(s). Below are the most recent \texttt{\{history\_length\}} observations and the corresponding actions you took: \texttt{\{action\_history\}}

You are now at step \texttt{\{current\_step\}} and your current observation is: \texttt{\{current\_observation\}}

Your admissible actions of the current situation are: [\texttt{\{admissible\_actions\}}].

Now it's your turn to take an action.

You should first reason step-by-step about the current situation. This reasoning process MUST be enclosed within \texttt{<think>} \texttt{</think>} tags.

Once you've finished your reasoning, you should choose an admissible action for current step and present it within \texttt{<action>} \texttt{</action>} tags.
\end{promptbox}

\begin{promptbox}{Hindsight Prompt Template for WebShop}
You are an expert autonomous agent operating in the WebShop e-commerce environment.

Your task is to: Navigate the website to reach the following exact target page state:

\texttt{[Target Page]: \{final\_obs\}}

Prior to this step, you have already taken \texttt{\{step\_count\}} step(s). Below are the most recent \texttt{\{history\_length\}} observations and the corresponding actions you took: \texttt{\{action\_history\}}

You are now at step \texttt{\{current\_step\}} and your current observation is: \texttt{\{current\_observation\}}.

Your admissible actions of the current situation are:
\[
\texttt{\{available\_actions\}}
\]

Now it's your turn to take one action for the current step.

You should first reason step-by-step about the current situation, then think carefully which admissible action best advances the shopping goal. This reasoning process MUST be enclosed within \texttt{<think>} \texttt{</think>} tags.

Once you've finished your reasoning, you should choose an admissible action for the current step and present it within \texttt{<action>} \texttt{</action>} tags.
\end{promptbox}

The previously executed action
\[
a_{i,t}=(y_{i,t,1},\ldots,y_{i,t,L_{i,t}})
\]
is then teacher-forced under this hindsight context to compute
$p^H_{i,t}$. The same action is teacher-forced under the original context
to compute $p_{i,t}$.

\section{Qualitative case studies}
\label{app:case_studies}

We provide qualitative examples illustrating how RELACE assigns
action-dependent retrospective credit. In each example, we report the
original-context score $p_{i,t}$, outcome-conditioned score $p^H_{i,t}$,
and the resulting trajectory-normalized relevance weight $\rho_{i,t}$.
These examples are intended to illustrate the behavior of the estimator,
rather than establish causal attribution.

\subsection{ALFWorld case study}
\label{app:alfworld_case}

We visualize a successful ALFWorld trajectory to illustrate the
multi-turn decision process in embodied environments. The agent receives
only a terminal success signal, while intermediate actions must be
assigned credit based on their contribution to completing the task.

\begin{promptbox}{ALFWorld Rollout: Step 1}

\textbf{Task:}

\texttt{cool some potato and put it in diningtable}

\vspace{0.5em}

\textbf{Observation:}

The agent is in the middle of a room. Available objects include a
diningtable, fridge, cabinets, drawers, microwave, sinkbasin, and other
kitchen objects.

\vspace{0.5em}

\textbf{Action:}

\texttt{go to diningtable 1}

\end{promptbox}

\begin{promptbox}{ALFWorld Rollout: Step 2}

\textbf{Observation:}

The agent arrives at the dining table.

Objects on the table include:

\texttt{apple 3, bread 1, butterknife 3, egg 2, fork 1,
knife 1, knife 2, mug 1, plate 1, potato 4, potato 2, tomato 3}

\vspace{0.5em}

\textbf{Action:}

\texttt{take potato 2 from diningtable 1}

\end{promptbox}

\begin{promptbox}{ALFWorld Rollout: Step 3}

\textbf{Observation:}

The agent is holding:

\texttt{potato 2}

Available actions include moving to the fridge.

\vspace{0.5em}

\textbf{Action:}

\texttt{go to fridge 1}

\end{promptbox}

\begin{promptbox}{ALFWorld Rollout: Step 4}

\textbf{Observation:}

The agent arrives at the fridge.

\texttt{The fridge 1 is closed.}

\vspace{0.5em}

\textbf{Action:}

\texttt{cool potato 2 with fridge 1}

\end{promptbox}

\begin{promptbox}{ALFWorld Rollout: Step 5}

\textbf{Observation:}

The agent successfully cools the potato:

\texttt{You cool the potato 2 using the fridge 1.}

\vspace{0.5em}

\textbf{Action:}

\texttt{go to diningtable 1}

\end{promptbox}

\begin{promptbox}{ALFWorld Rollout: Step 6}

\textbf{Observation:}

The agent returns to the dining table with the cooled potato.

\vspace{0.5em}

\textbf{Action:}

\texttt{move potato 2 to diningtable 1}

\vspace{0.5em}

\textbf{Outcome:}

\texttt{success}

\end{promptbox}

\subsection{WebShop case study}
\label{app:case_webshop}

We show a successful WebShop trajectory to illustrate the multi-turn
interaction structure used by our agent. The trajectory receives a single
terminal success reward, while RELACE assigns credit at the action level.

\begin{promptbox}{WebShop Rollout: Step 1}

\textbf{Task:}

Find me machine wash men's tuxedo shirts with polyester heathers,
heathers cotton, cotton heather, needle sleeve, classic fit with
color: heather grey, fit type: youth, size: medium, and price lower
than 30.00 dollars.

\vspace{0.5em}

\textbf{Observation:}

\texttt{Search}

\vspace{0.5em}

\textbf{Action:}

\texttt{search[machine wash men's tuxedo shirts | polyester heathers |
heathers cotton | cotton heather | needle sleeve | classic fit |
color: heather grey | fit type: youth | size: medium | price < 30.00]}

\end{promptbox}

\begin{promptbox}{WebShop Rollout: Step 2}

\textbf{Observation:}

\texttt{Page 1 (Total results: 50)}

\texttt{B07JVVDJ6L -- Canadian Tuxedo T Shirt - Novelty Denim Tux-T-Shirt}

\texttt{\$19.99}

\vspace{0.5em}

\textbf{Action:}

\texttt{click[B07JVVDJ6L]}

\end{promptbox}

\begin{promptbox}{WebShop Rollout: Step 3}

\textbf{Observation:}

Product page:

\texttt{fit type: men, women, youth}

\texttt{color: royal blue, baby blue, heather grey, heather blue}

\texttt{small, medium, large, x-large, xx-large, 3x-large,
2t, 3t, 4t, x-small}

\vspace{0.5em}

\textbf{Action:}

\texttt{click[youth]}

\end{promptbox}

\begin{promptbox}{WebShop Rollout: Step 4}

\textbf{Observation:}

Product page after selecting fit type:

\texttt{fit type: youth}

Available colors:

\texttt{royal blue, baby blue, heather grey, heather blue}

\vspace{0.5em}

\textbf{Action:}

\texttt{click[heather grey]}

\end{promptbox}

\begin{promptbox}{WebShop Rollout: Step 5}

\textbf{Observation:}

Product page after selecting color:

\texttt{color: heather grey}

Available sizes:

\texttt{small, medium, large, x-large, xx-large}

\vspace{0.5em}

\textbf{Action:}

\texttt{click[medium]}

\end{promptbox}

\begin{promptbox}{WebShop Rollout: Step 6}

\textbf{Observation:}

Final product configuration:

\texttt{Product: Canadian Tuxedo T Shirt - Novelty Denim Tux-T-Shirt}

\texttt{color: heather grey}

\texttt{fit type: youth}

\texttt{size: medium}

\texttt{price: \$19.99}

\vspace{0.5em}

\textbf{Action:}

\texttt{click[buy now]}

\vspace{0.5em}

\textbf{Outcome:}

\texttt{Successful purchase}

\end{promptbox}


\begin{thebibliography}{22}
\providecommand{\natexlab}[1]{#1}
\providecommand{\url}[1]{\texttt{#1}}
\expandafter\ifx\csname urlstyle\endcsname\relax
  \providecommand{\doi}[1]{doi: #1}\else
  \providecommand{\doi}{doi: \begingroup \urlstyle{rm}\Url}\fi

\bibitem[Ahmadian et~al.(2024)Ahmadian, Cremer, Gall{\'e}, Fadaee, Kreutzer,
  Pietquin, {\"U}st{\"u}n, and Hooker]{ahmadian2024rloo}
Arash Ahmadian, Chris Cremer, Matthias Gall{\'e}, Marzieh Fadaee, Julia
  Kreutzer, Olivier Pietquin, Ahmet {\"U}st{\"u}n, and Sara Hooker.
\newblock Back to basics: Revisiting {REINFORCE}-style optimization for
  learning from human feedback in {LLMs}.
\newblock In \emph{Proceedings of the 62nd Annual Meeting of the Association
  for Computational Linguistics (Volume 1: Long Papers)}, 2024.
\newblock URL \url{https://arxiv.org/abs/2402.14740}.

\bibitem[Andrychowicz et~al.(2017)Andrychowicz, Wolski, Ray, Schneider, Fong,
  Welinder, McGrew, Tobin, Abbeel, and Zaremba]{andrychowicz2017hindsight}
Marcin Andrychowicz, Filip Wolski, Alex Ray, Jonas Schneider, Rachel Fong,
  Peter Welinder, Bob McGrew, Josh Tobin, Pieter Abbeel, and Wojciech Zaremba.
\newblock Hindsight experience replay.
\newblock In \emph{Advances in Neural Information Processing Systems},
  volume~30, pp.\  5048--5058, 2017.
\newblock URL
  \url{https://proceedings.neurips.cc/paper/2017/hash/453fadbd8a1a3af50a9df4df899537b5-Abstract.html}.

\bibitem[Cheng et~al.(2026)Cheng, He, Feng, Xu, Yan, Feng, and
  An]{cheng2026graphgpo}
Xin Cheng, Shuo He, Lang Feng, HaiYang Xu, Ming Yan, Lei Feng, and Bo~An.
\newblock Beyond trajectory-level attribution: Graph-based credit assignment
  for agentic reinforcement learning.
\newblock \emph{arXiv preprint arXiv:2605.26684}, 2026.
\newblock URL \url{https://arxiv.org/abs/2605.26684}.

\bibitem[Feng et~al.(2025)Feng, Xue, Liu, and An]{feng2025gigpo}
Lang Feng, Zhenghai Xue, Tingcong Liu, and Bo~An.
\newblock Group-in-group policy optimization for {LLM} agent training.
\newblock In \emph{Advances in Neural Information Processing Systems},
  volume~38, 2025.
\newblock \doi{10.52202/085713-1544}.
\newblock URL \url{https://arxiv.org/abs/2505.10978}.

\bibitem[Harutyunyan et~al.(2019)Harutyunyan, Dabney, Mesnard, Azar, Piot,
  Heess, van Hasselt, Wayne, Singh, Precup, and
  Munos]{harutyunyan2019hindsight}
Anna Harutyunyan, Will Dabney, Thomas Mesnard, Mohammad~Gheshlaghi Azar, Bilal
  Piot, Nicolas Heess, Hado~P. van Hasselt, Gregory Wayne, Satinder Singh,
  Doina Precup, and Remi Munos.
\newblock Hindsight credit assignment.
\newblock In \emph{Advances in Neural Information Processing Systems},
  volume~32, pp.\  12467--12476, 2019.
\newblock URL
  \url{https://proceedings.neurips.cc/paper_files/paper/2019/hash/195f15384c2a79cedf293e4a847ce85c-Abstract.html}.

\bibitem[He et~al.(2026)He, Feng, Wei, Cheng, Feng, and An]{hgpo2026}
Shuo He, Lang Feng, Qi~Wei, Xin Cheng, Lei Feng, and Bo~An.
\newblock Hierarchy-of-groups policy optimization for long-horizon agentic
  tasks.
\newblock In \emph{International Conference on Learning Representations}, 2026.
\newblock URL \url{https://arxiv.org/abs/2602.22817}.

\bibitem[Li et~al.(2026)Li, Zhou, Duan, and Chen]{li2026ecpo}
Yuanfan Li, Qi~Zhou, Wenjing Duan, and Lu~Chen.
\newblock When denser credit is not enough: Evidence-calibrated policy
  optimization for long-horizon llm agent training.
\newblock \emph{arXiv preprint arXiv:2606.05885}, 2026.
\newblock \doi{10.48550/arXiv.2606.05885}.
\newblock URL \url{https://arxiv.org/abs/2606.05885}.

\bibitem[Lightman et~al.(2023)Lightman, Kosaraju, Burda, Edwards, Baker, Lee,
  Leike, Schulman, Sutskever, and Cobbe]{lightman2023verify}
Hunter Lightman, Vineet Kosaraju, Yura Burda, Harri Edwards, Bowen Baker, Teddy
  Lee, Jan Leike, John Schulman, Ilya Sutskever, and Karl Cobbe.
\newblock Let's verify step by step.
\newblock \emph{arXiv preprint arXiv:2305.20050}, 2023.
\newblock URL \url{https://arxiv.org/abs/2305.20050}.

\bibitem[Meulemans et~al.(2023)Meulemans, Schug, Kobayashi, Daw, and
  Wayne]{meulemans2023cocoa}
Alexander Meulemans, Simon Schug, Seijin Kobayashi, Nathaniel Daw, and Gregory
  Wayne.
\newblock Would i have gotten that reward? long-term credit assignment by
  counterfactual contribution analysis.
\newblock In \emph{Advances in Neural Information Processing Systems}, 2023.
\newblock URL
  \url{https://papers.neurips.cc/paper_files/paper/2023/hash/d8bd445c2abe1343cce0e14b361b2fb3-Abstract-Conference.html}.

\bibitem[{Qwen Team}(2024)]{qwen2024}
{Qwen Team}.
\newblock {Qwen2.5} technical report.
\newblock \emph{arXiv preprint arXiv:2412.15115}, 2024.
\newblock URL \url{https://arxiv.org/abs/2412.15115}.

\bibitem[Rauber et~al.(2019)Rauber, Ummadisingu, Mutz, and
  Schmidhuber]{rauber2019hpg}
Paulo Rauber, Avinash Ummadisingu, Filipe Mutz, and J{\"u}rgen Schmidhuber.
\newblock Hindsight policy gradients.
\newblock In \emph{International Conference on Learning Representations}, 2019.
\newblock URL \url{https://arxiv.org/abs/1711.06006}.

\bibitem[Schulman et~al.(2017)Schulman, Wolski, Dhariwal, Radford, and
  Klimov]{schulman2017ppo}
John Schulman, Filip Wolski, Prafulla Dhariwal, Alec Radford, and Oleg Klimov.
\newblock Proximal policy optimization algorithms.
\newblock \emph{arXiv preprint arXiv:1707.06347}, 2017.
\newblock \doi{10.48550/arXiv.1707.06347}.
\newblock URL \url{https://arxiv.org/abs/1707.06347}.

\bibitem[Shao et~al.(2024)Shao, Wang, Zhu, Xu, Song, Bi, Zhang, Zhang, Li, Wu,
  and Guo]{shao2024deepseekmath}
Zhihong Shao, Peiyi Wang, Qihao Zhu, Runxin Xu, Junxiao Song, Xiao Bi, Haowei
  Zhang, Mingchuan Zhang, Y.~K. Li, Y.~Wu, and Daya Guo.
\newblock {DeepSeekMath}: Pushing the limits of mathematical reasoning in open
  language models.
\newblock \emph{arXiv preprint arXiv:2402.03300}, 2024.
\newblock URL \url{https://arxiv.org/abs/2402.03300}.

\bibitem[Shinn et~al.(2023)Shinn, Cassano, Berman, Gopinath, Narasimhan, and
  Yao]{shinn2023reflexion}
Noah Shinn, Federico Cassano, Edward Berman, Ashwin Gopinath, Karthik
  Narasimhan, and Shunyu Yao.
\newblock Reflexion: Language agents with verbal reinforcement learning.
\newblock In \emph{Advances in Neural Information Processing Systems}, 2023.
\newblock URL \url{https://arxiv.org/abs/2303.11366}.

\bibitem[Shridhar et~al.(2021)Shridhar, Yuan, C{\^o}t{\'e}, Bisk, Trischler,
  and Hausknecht]{shridhar2021alfworld}
Mohit Shridhar, Xingdi Yuan, Marc-Alexandre C{\^o}t{\'e}, Yonatan Bisk, Adam
  Trischler, and Matthew Hausknecht.
\newblock {ALFWorld}: Aligning text and embodied environments for interactive
  learning.
\newblock In \emph{International Conference on Learning Representations}, 2021.
\newblock URL \url{https://arxiv.org/abs/2010.03768}.

\bibitem[Tan et~al.(2026)Tan, Yang, Chen, Shao, Wen, Shen, Luo, Du, Guo, and
  Li]{tan2026hcapo}
Hui-Ze Tan, Xiao-Wen Yang, Hao Chen, Jie-Jing Shao, Yi~Wen, Yuteng Shen,
  Weihong Luo, Xiku Du, Lan-Zhe Guo, and Yu-Feng Li.
\newblock Hindsight credit assignment for long-horizon {LLM} agents.
\newblock \emph{arXiv preprint arXiv:2603.08754}, 2026.
\newblock \doi{10.48550/arXiv.2603.08754}.
\newblock URL \url{https://arxiv.org/abs/2603.08754}.

\bibitem[Wang et~al.(2025{\natexlab{a}})Wang, Liu, Fu, Li, Wang, Lin, Yue,
  Zhang, Wang, and Wang]{wang2025empg}
Jiawei Wang, Jiacai Liu, Yuqian Fu, Yingru Li, Xintao Wang, Yuan Lin, Yu~Yue,
  Lin Zhang, Yang Wang, and Ke~Wang.
\newblock Harnessing uncertainty: Entropy-modulated policy gradients for
  long-horizon {LLM} agents.
\newblock \emph{arXiv preprint arXiv:2509.09265}, 2025{\natexlab{a}}.
\newblock URL \url{https://arxiv.org/abs/2509.09265}.

\bibitem[Wang(2026)]{wang2026darkroom}
Yu~Wang.
\newblock The dark room in the reward channel: Dense prediction rewards
  collapse {GRPO}-trained {LLM} agents -- and the channel, not the content,
  decides what works.
\newblock \emph{arXiv preprint arXiv:2607.21273}, 2026.
\newblock \doi{10.48550/arXiv.2607.21273}.
\newblock URL \url{https://arxiv.org/abs/2607.21273}.

\bibitem[Wang et~al.(2025{\natexlab{b}})Wang, Wang, Wang, Zhang, Li, Yang, Jin,
  Yu, Nguyen, Liu, Gottlieb, Lu, Cho, Wu, Fei-Fei, Wang, Choi, and
  Li]{wang2025ragen}
Zihan Wang, Kangrui Wang, Qineng Wang, Pingyue Zhang, Linjie Li, Zhengyuan
  Yang, Xing Jin, Kefan Yu, Minh~Nhat Nguyen, Licheng Liu, Eli Gottlieb, Yiping
  Lu, Kyunghyun Cho, Jiajun Wu, Li~Fei-Fei, Lijuan Wang, Yejin Choi, and
  Manling Li.
\newblock {RAGEN}: Understanding self-evolution in {LLM} agents via multi-turn
  reinforcement learning.
\newblock \emph{arXiv preprint arXiv:2504.20073}, 2025{\natexlab{b}}.
\newblock URL \url{https://arxiv.org/abs/2504.20073}.

\bibitem[Xu et~al.(2025)Xu, Zhao, Wang, Li, Jin, Jin, Mei, Wan, and
  Metaxas]{xu2025epo}
Wujiang Xu, Wentian Zhao, Zhenting Wang, Yu-Jhe Li, Can Jin, Mingyu Jin, Kai
  Mei, Kun Wan, and Dimitris~N. Metaxas.
\newblock {EPO}: Entropy-regularized policy optimization for {LLM} agents
  reinforcement learning.
\newblock \emph{arXiv preprint arXiv:2509.22576}, 2025.
\newblock URL \url{https://arxiv.org/abs/2509.22576}.

\bibitem[Yao et~al.(2022)Yao, Chen, Yang, and Narasimhan]{yao2022webshop}
Shunyu Yao, Howard Chen, John Yang, and Karthik Narasimhan.
\newblock {WebShop}: Towards scalable real-world web interaction with grounded
  language agents.
\newblock In \emph{Advances in Neural Information Processing Systems}, 2022.
\newblock URL \url{https://arxiv.org/abs/2207.01206}.

\bibitem[Yao et~al.(2023)Yao, Zhao, Yu, Du, Shafran, Narasimhan, and
  Cao]{yao2023react}
Shunyu Yao, Jeffrey Zhao, Dian Yu, Nan Du, Izhak Shafran, Karthik Narasimhan,
  and Yuan Cao.
\newblock {ReAct}: Synergizing reasoning and acting in language models.
\newblock In \emph{International Conference on Learning Representations}, 2023.
\newblock URL \url{https://arxiv.org/abs/2210.03629}.

\end{thebibliography}
\end{document}